\pdfoutput=1
\documentclass[11pt]{article}

\usepackage[]{emnlp2021}

\usepackage{times}
\usepackage{latexsym}

\usepackage[T1]{fontenc}
\usepackage[utf8]{inputenc}

\usepackage{graphicx}
\usepackage{tabularx}
\usepackage{wrapfig}
\usepackage[normalem]{ulem}
\usepackage{amsmath}
\usepackage{subcaption}
\usepackage{color,colortbl}
\usepackage{multirow}
\usepackage{cancel}
\usepackage{multicol}
\usepackage{makecell}
\usepackage{comment}
\usepackage{amssymb}
\usepackage{ulem}
\usepackage{dcolumn}
\usepackage{url}
\usepackage{xcolor} 
\usepackage{paralist}
\usepackage{inconsolata}
\usepackage{rotating}

\usepackage{microtype}

\newcommand{\aspect}[1]{\textsf{\fontsize{9pt}{.1pt}\selectfont{#1}}}

\newcommand{\keyword}[1]{\textrm{\fontsize{11pt}{.2pt}\selectfont{#1}}}

\newcommand{\mc}{\multicolumn}
\newcolumntype{d}[1]{D{.}{.}{#1}}

\definecolor{myorange}{rgb}{0.93,0.49,0.2}
\definecolor{mygreen}{rgb}{0.33,0.51,0.21}
\definecolor{myblue}{rgb}{0.18,0.46,0.71}

\makeatletter
\newcommand{\thickhline}{%
	\noalign {\ifnum 0=`}\fi \hrule height 1pt
	\futurelet \reserved@a \@xhline
}
\makeatother

\title{Aspect-Controllable Opinion Summarization}

\author{
	Reinald Kim Amplayo \quad
	Stefanos Angelidis \quad
	Mirella Lapata \\
	Institute for Language, Cognition and Computation \\
	School of Informatics, University of Edinburgh \\
10 Crichton Street, EH8 9AB \\
	\texttt{reinald.kim@ed.ac.uk\quad s.angelidis@ed.ac.uk\quad mlap@inf.ed.ac.uk}
}

\begin{document}
\maketitle
\begin{abstract}

  Recent work on opinion summarization produces general summaries
  based on a set of input reviews and the popularity of opinions
  expressed in them.
%  relevance. However, the quality of summaries is dependent on user
%  needs. For example, one customer may find the image quality of a
%  television the most important aspect, while another might only care
%  about the connectivity.  
  In this paper, we propose an approach that allows the generation of
  customized summaries based on aspect queries (e.g.,~describing the
  location and room of a hotel).  Using a review
  corpus, we create a synthetic training dataset of (review,
  summary) pairs enriched with \textit{aspect controllers} which are
  induced by a multi-instance learning model that predicts the aspects
  of a document at different levels of granularity.  We fine-tune a
  pretrained model using our synthetic dataset and generate
  aspect-specific summaries by modifying the aspect controllers.
  Experiments on two benchmarks show that our model outperforms the
  previous state of the art and generates personalized summaries by
  controlling the number of aspects discussed in them.
\end{abstract}

\section{Introduction}

%\begin{figure}[t]
%	\centering
%	\includegraphics[width=\columnwidth]{../intro2}
%	\caption{Aspect-controlled opinion summarization: given a set
%          of reviews about an entity (e.g., television reviews) and an
%          aspect query (e.g., sound and image quality), generate a
%          summary that contains opinions regarding the target
%          aspects. Our approach is to induce three aspect controllers:
%          aspect-related keywords, review sentences, and aspect codes,
%          and use them to guide the summary generation towards correct
%          aspects.}
%\label{fig:intro}
%\end{figure}

Consumers oftentimes resort to review websites to inform their
decision making (e.g.,~whether to buy a product or use a service).
The proliferation of online reviews has accelerated research on
opinion mining \cite{pang2007opinion}, where the ultimate goal is to
glean information from reviews so that users can make decisions more
efficiently.  Opinion mining has assumed several guises in the
literature such as sentiment analysis \cite{pang2002thumbs}, aspect
extraction \cite{hu-liu-2004-mining, he2017unsupervised}, combinations
thereof \cite{mukherjee2012aspect,pontiki2016semeval}, and notably
\textit{opinion summarization} \cite{hu2006opinion,wang2016neural},
whose aim is to create a textual summary of opinions found in multiple
reviews.

Text summarization models, both extractive
\cite{narayan2018ranking,zheng2019sentence,cachola2020tldr} and
abstractive \cite{see2017get,gehrmann2018end,liu2019text}, operate
under the assumption that salient content is relevant
\cite{erkan2004lexrank} and should be presented in the
summary. Opinion summarization is no exception, focusing on creating
summaries based on opinions that are \textit{popular} or
\textit{redundant} across reviews
\cite{angelidis2018summarizing,chu2019meansum,amplayo2020unsupervised,bravzinskas2019unsupervised,amplayo2021unsupervised}.

% usually operate following the notion of
%salience in documents to find relevant content
%\cite{erkan2004lexrank}; extractive models
%\cite{narayan2018ranking,zheng2019sentence,cachola2020tldr} selects
%salient sentences as summaries, while abstractive models
% generate summaries
%focusing on salient information. Indeed, current opinion summarization
%models follow this idea by generating summaries based on opinions
%about aspects (e.g., image and sound quality of televisions) that are
%considered \textit{popular} or \textit{redundant} across different
%reviews

\begin{table}[t]
\small
\centering
\begin{tabular}{@{~}p{7.5cm}@{~}}
  \thickhline
  \textbf{\aspect{General}}\\
  The room was clean and comfortable. The staff was very friendly and
  helpful. It was a great location, just a short walk to the
  beach. There wasn't much to do in the area, but the food was
  good.\\ \hline
  \textbf{\aspect{\textcolor{red}{Location}}}  \\\textcolor{red}{The location was great, right on the Boardwalk, and close to the
    Venice beach.}\\\hline
  \textbf{\aspect{\textcolor{blue}{Rooms}}} \\ \textcolor{blue}{The room was very clean and the bathroom was very nice. The bathroom had a large separate shower. There was a TV in the room.}\\
  \hline
  \textbf{\aspect{\textcolor{red}{Location}}} and
  \textbf{\aspect{\textcolor{blue}{Rooms}}} \\ \textcolor{red}{The location is great, right on Boardwalk, and the beach is very nice.} \textcolor{blue}{The room was very clean and the bathroom was very nice and the shower was great.}\\
  \hline
  \textbf{\aspect{\textcolor{cyan}{Cleanliness}}}, \textbf{\aspect{\textcolor{red}{Location}}}, \textbf{\aspect{\textcolor{blue}{Room}}}, and \textbf{\aspect{\textcolor{orange}{Service}}}\\ \textcolor{orange}{The staff was very friendly and helpful.} \textcolor{cyan}{The room was very clean, and the bathroom was very nice.} \textcolor{red}{It was a great location, right on the beach.}
  \\
  \thickhline
\end{tabular}
\vspace*{-1ex}
\caption{General and aspect-specific summaries generated by our model
  for a hotel from the \textsc{Space} dataset. Aspects and
  aspect-specific sentences are color-coded.}
\label{tab:examples}%
\end{table}

However, the notion of salience in reviews largely depends on
\textit{user interest}. For example, one might only care about the
connectivity of a television product, an aspect which might be
unpopular amongst reviews.  As a result, models that create
\emph{general} opinion summaries may not satisfy the needs of all
users, limiting their ability to make decisions.
\citet{angelidis2020extractive} mitigate this problem with an
extractive approach that produces both general and
\textit{aspect-specific} opinion summaries. They achieve this
essentially by clustering opinions through a discrete latent variable
model \cite{vandenoord2017neural} and extracting sentences based on
popular aspects or a particular aspect. By virtue of being extractive,
their summaries can be incoherent, and verbose containing unnecessary
redundancy. And although their model creates summaries for individual
aspects, it is not clear how to control the number of aspects in the
output (e.g.,~to obtain summaries that mention multiple rather than a
single aspect of an entity).

In this paper, we propose an abstractive opinion summarization model
that generates aspect-controllable summaries. Using a corpus of
reviews on entities (e.g.,~hotels, television sets), we construct a
synthetic training dataset consisting of reviews, a pseudo-summary,
and three types of \emph{aspect controllers} which reflect different
levels of granularity: aspect-related keywords, review sentences, and
document-level aspect codes.  We induce aspect controllers
automatically based on a multiple instance learning model
\cite{keeler1991self} and very little human involvement. Using the
aspect-enriched dataset, we then fine-tune a pretrained model
\cite{raffel2020exploring} on summary generation.  By modifying the
controllers, we can flexibly generate general and aspect-specific
summaries, discussing one or more aspects. Figure~\ref{tab:examples}
shows summaries generated by our model.

%Our model is purely
%unsupervised, i.e., we do not rely on gold-standard summaries.

We perform experiments on \textsc{Space}
\cite{angelidis2020extractive}, a single domain dataset consisting of
hotel reviews, and \textsc{Oposum} \cite{angelidis2018summarizing}, a
dataset with product reviews from multiple domains (e.g.,~``laptop
bags'', ``boots''). Automatic and human evaluation show that our
model outperforms previous approaches on both tasks of general and
aspect-specific summarization. We also demonstrate that it can
effectively generate multi-aspect summaries based on user preferences.
We make our code and data publicly available.\footnote{
\url{https://github.com/rktamplayo/AceSum}}

\section{Related Work}

Earlier work on opinion summarization has focused on general
summarization using extractive
\cite{hu2006opinion,kim2011comprehensive,angelidis2018summarizing} or
abstractive methods
\cite{ganesan2010opinosis,carenini2013multi,difabbrizio2014hybrid}.
Due to the absence of opinion summaries in review websites and the
difficulty of annotating them on a large scale, more recent methods
consider an unsupervised learning setting where there are only reviews
available without corresponding summaries
\cite{chu2019meansum,bravzinskas2019unsupervised}.  They make use of
autoencoders \cite{kingma2014auto} and variants thereof to learn a
review decoder through reconstruction, and use it to generate
summaries conditioned on averaged representations of the inputs.

A more successful approach to opinion summarization is through the
creation of synthetic datasets, where (review, summary) pairs are
constructed from a review corpus to enable supervised training. These
methods usually start by randomly selecting a review which they treat
as a pseudo-summary and subsequently pair it with a set of input
reviews based on different strategies. These include random sampling
\cite{bravzinskas2019unsupervised}, generating noisy versions of the
pseudo-summary \cite{amplayo2020unsupervised}, ranking reviews based
on similarity and relevance \cite{elsahar2020self}, and making use of
content plans to create more naturalistic pairs
\cite{amplayo2021unsupervised}.

Our work is closest to \citet{angelidis2020extractive} who propose
an extractive summarization model that uses a vector-quantized
variational autoencoder \cite{vandenoord2017neural} to learn
aspect-specific review representations. Their model effectively groups
opinion sentences into clusters and extracts those capturing
aspect-relevant information. We employ multi-instance learning to
identify aspect-bearing elements in reviews with varying degrees of
granularity (e.g.,~words, sentences, documents) which we argue affords
greater flexibility and better control of the output summaries. In
doing so, we also introduce an effective method to create synthetic
datasets for aspect-guided opinion summarization.  Our work also
relates to approaches which attempt to control summarization output
based on length \cite{kikuchi2016controlling}, content
\cite{fan2018controllable}, style \cite{cao2021inference}, or textual
queries \cite{dang2006duc}. Although we focus solely on aspect, our
method is general and could be used to adjust additional properties of
a summary such as sentiment (e.g.,~positive vs. negative) or style
(e.g.,~formal vs. colloquial).

\begin{figure*}[t]
	\centering
	\includegraphics[width=\textwidth]{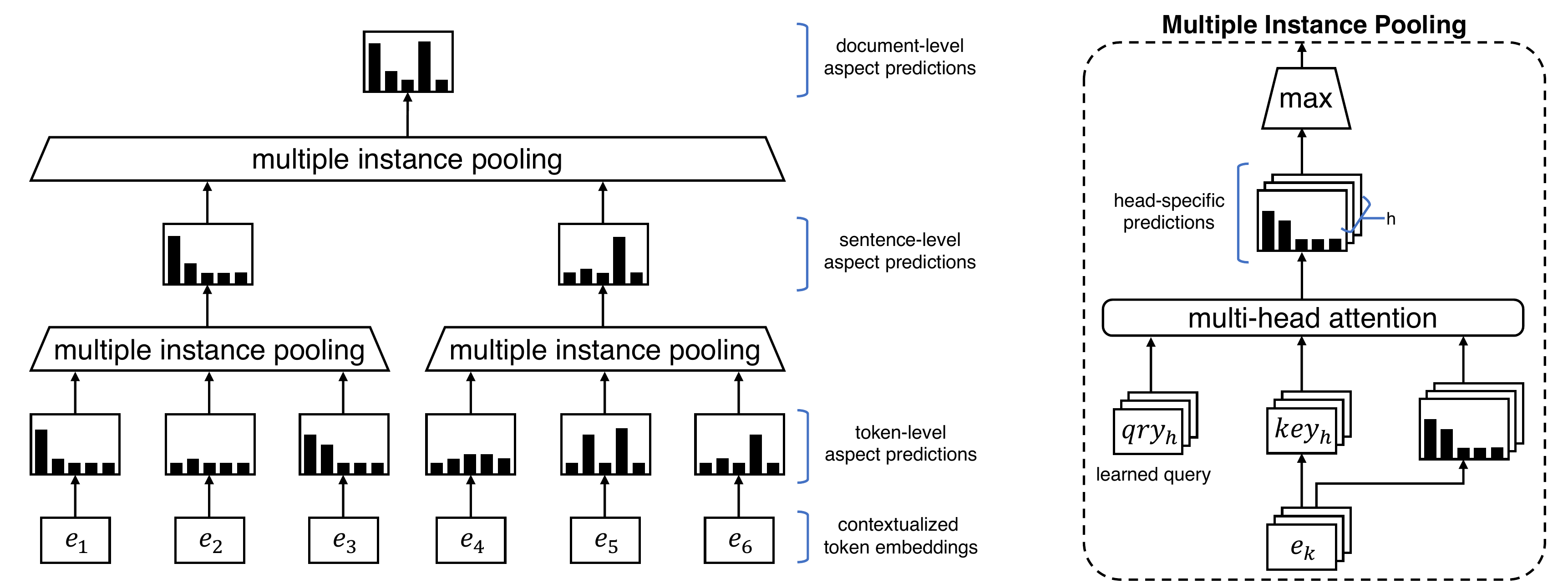}
\vspace*{-3ex}
	\caption{Overview of the controller induction
          model. Token-level aspect predictions are aggregated into 
          sentence-level predictions using a multiple instance
          pooling mechanism (described on the right). The process is
          repeated from sentence- to document-level predictions.}
	\label{fig:mil-model}
\end{figure*}

\section{Problem Formulation}

Let~$C$ denote a corpus of reviews about entities (e.g.,~products,
hotels). Let \mbox{$R_e=\{r_1,r_2,...,r_N\}$} denote a set of reviews
for entity $e$ and \mbox{$A_e=\{a_1,a_2,...,a_M\}$} a set of aspects
that are relevant for the entity (e.g.,~\aspect{cleanliness},
\aspect{location}). Each review~$r_i$ is a sequence of tokens
$\{w_1,w_2,...\}$, while each aspect $a_j$ is represented by a small
set of \textit{seed words} $\{v_1,v_2,...\}$ (e.g., \textsl{spotless},
\textsl{dirty, stain}). These seed words can be acquired automatically
\cite{angelidis2018summarizing} or provided by users (see Appendix for
those used in our experiments).

Our approach creates two types of summaries: (a) a general summary
that contains salient opinions about \textit{all} aspects of an
entity, and (b) an aspect-specific summary that focuses on opinions
about \emph{particular} aspects of interest specified by a query
$Q=\{q_1,q_2,...,q_M\}$; here,~$q_j$ is an indicator function which
designates whether the aspect should be mentioned in the summary. We
emphasize that the query can represent more than one aspect to reflect
real-world usage.  To facilitate supervised training, we create a
synthetic training dataset $D={(X, z, y)}$, which is a set of triples
composed of input reviews~$X$, a pseudo-summary~$y$, and aspect
controllers~$z$ (Section~\ref{sec:dataset-creation}).  Our aspect
controllers are induced with a unified model based on multi-instance
learning (Section~\ref{sec:controller-induction})  and correspond to
different levels of granularity: (1) document-level aspect codes,
(2)~aspect-related review sentences, and (3)~aspect keywords.

At training time, we fine-tune a pretrained sequence-to-sequence
Transformer model \cite{raffel2020exploring} using controllers $z$ as
input and a pseudo-summary as output. During inference, we modulate
summary generation by modifying the controllers, e.g.,~we produce a
general summary using all aspect codes, or an aspect-specific one
based on a subset thereof (Section~\ref{sec:opinion-summarization}).

\subsection{Controller Induction Model}
\label{sec:controller-induction}

A key feature of our approach is the set of aspect controllers which allow
our summarization model to be controllable.
%We use three types of aspect controllers that corresponds to different levels of granularity: (1) keywords that are related to aspects mentioned in the summary, (2) review sentences that mention similar aspects in the summary, and (3) a document-level aspect switch that represents the aspects found in the summary. 
We induce these controllers using a multiple instance learning (MIL)
model, illustrated in Figure~\ref{fig:mil-model}. MIL is a machine
learning framework where labels are associated with groups of
instances (i.e., \textit{bags}), while instance labels are unobserved
\cite{keeler1991self}.  The goal is then to infer labels for bags
\cite{dietterich1997solving, maron1998multiple} or
jointly for instances and bags
\cite{zhou2009multi,wei2014scalable,kotzias2015group,xu2019weakly,angelidis2018multiple}. Our
MIL model is an example of the latter variant.

In our setting, documents are bags of sentences and sentences are bags
of tokens. We further assume that only documents have aspect labels.
Given review $r$ with tokens~$\{w_k\}$, we obtain token encodings
$\mathbf{e} = \{e_k\}$ from a pretrained language model (PLM;
\citealt{liu2019roberta}) which uses the popular Transformer
architecture \cite{vaswani2017attention}.  We use a non-linear
transformation to obtain token-level aspect predictions
$\mathbf{z}_{\mathcal{T}}$:
\begin{align}
	\mathbf{e} &= \operatorname{PLM}(\{w_k\}) \\ \label{eq:1}
	\mathbf{z}_{\mathcal{T}} &= \tanh (W \mathbf{e} + b) 
\end{align}
where $\mathbf{z}_{\mathcal{T}} \in \mathbb{R}^{N \times M}$, and $N$
and $M$ are the number of tokens and aspects, respectively. A positive
value denotes that the token is related to the aspect of interest (and
otherwise unrelated).

\paragraph{Multiple Instance Pooling}

To obtain sentence-level aspect predictions
$\mathbf{z}_{\mathcal{S}}$, we aggregate token-level predictions
$\mathbf{z}_{\mathcal{T}}$ using a new pooling method particularly
effective for our multi-instance learning setting. We first obtain
multiple predictions $\mathbf{z}_h$ for each attention head~$h$:
\begin{align}
	\mathbf{z}_h &= \sum\nolimits_k(\mathbf{z}_{\mathcal{T}} * a_h[k])\\
	a_h &= \text{softmax}(key_h \cdot qry_h) 
\end{align}
where $*$ is element-wise multiplication, $\cdot$ is dot product, $k$ is the token index, $qry_h$ is a head-specific query vector, and $key_h$ is defined below:
\begin{align}
key_h &= \tanh(W_{h} \mathbf{e} + b_{h}) 
\end{align}
We hypothesize that different attention heads represent different
aspects of the semantic space, and are thus helpful at predicting
multiple aspects.  We obtain a sentence-level prediction by max
pooling the predictions of individual heads:
\begin{align}
\mathbf{z}_{S} &= \text{max-pool}(\{\mathbf{z}_h\})
\end{align}

We use max pooling since we want to isolate the most pertinent aspects
for a given sentence; standard pooling methods such as mean and
attention pooling \cite{angelidis2018multiple,xu2019weakly} assume
that \emph{all} instances of a bag contribute to its label.  In
Figure~\ref{fig:mil-model} (right) we illustrate our pooling mechanism
and empirically show in experiments (see
Section~\ref{sec:auto-results}) it is superior to alternatives.

% using mean pooling to aggregate head predictions would lower the
% confidence of the model at labelling the sentence with the aspect
% \aspect{service}.  We empirically show (see
% Section~\ref{sec:auto-results}) that max-pooling is superior to
% alternative pooling strategies.

We so far discussed how multiple instance pooling is applied at the
token-level to obtain sentence-level predictions
$\mathbf{z}_{\mathcal{S}}$. Analogously, multiple instance pooling is
applied to sentences to obtain document-level predictions
$\mathbf{z}_{\mathcal{D}}$ (see Figure~\ref{fig:mil-model}).

%= \text{MIP}(\mathbf{z}_{tok})$, as well as at the sentence-level to
%obtain the document-level prediction $\mathbf{z}_{doc}
%= \text{MIP}(\mathbf{z}_{sen})$.

\paragraph{Training and Inference}

Training the multiple instance model just described requires a dataset
consisting of (review, aspect label) pairs. Unfortunately, we do not
have access to annotations denoting which aspects are discussed in
each review. Recall, however, that aspects are represented by seed
words $\{v_1,v_2,...\}$, which we exploit to induce silver-standard
labels. Specifically, for each review in the dataset, we obtain binary
labels $\mathbf{\hat{z}}_{\mathcal{D}}$ where
$\mathbf{\hat{z}}_{\mathcal{D}}[a] = 1$ if at least one seed word for
aspect~$a$ is found in the review (and $-1$ otherwise).

We train the model using a soft margin loss, summing over all aspects
$a\in A$:
\begin{equation}
	\mathcal{L}_{ctrl} = \sum\nolimits_a \log(1
	+ \exp(-\mathbf{z}_{\mathcal{D}}[a]*\mathbf{\hat{z}}_{\mathcal{D}}[a])) \label{eq:softmargin}
\end{equation}
The parameters of the pretrained language model (see
Equation~\eqref{eq:1}) are frozen, i.e.,~they are not fine-tuned
during training which makes our controller induction model lightweight
and efficient.

%\begin{figure}[t]
%	\centering
%	\includegraphics[width=\columnwidth]{../controller-induction}
%	\caption{Illustration of how the aspect controllers are
%          induced. The aspect code is first induced from the
%          summary. Then, aspect keywords and sentences are induced
%          from the reviews, which are then ranked and selected using
%          the given aspect code. At test time, we generate different
%          kinds of summaries by modifying aspect codes based on user
%          request.  }
%	\label{fig:controller-induction}
%\end{figure}

\newcommand\bleuline{\bgroup\markoverwith{\textcolor{blue}{\rule[-0.5ex]{2pt}{0.4pt}}}\ULon}
\definecolor{Gray}{gray}{0.9}
\definecolor{LightCyan}{rgb}{0.88,1,1}

\begin{figure}[t]
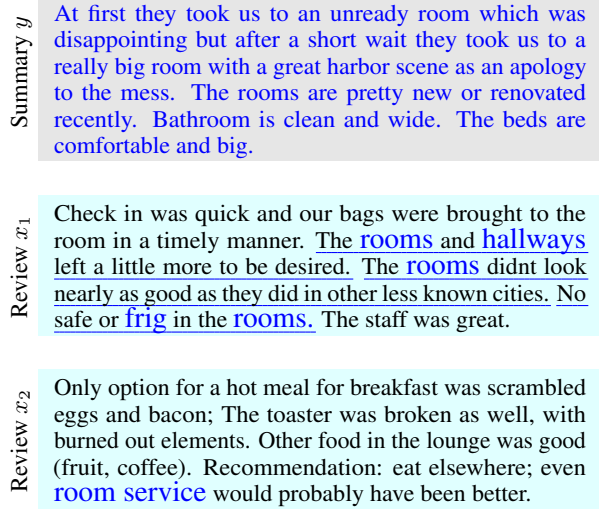

\begin{small}
	\setlength{\extrarowheight}{0.7ex}%
\hspace*{-1ex}\begin{tabular}{@{}l@{\hspace{2ex}}p{7cm}@{}}
  \raisebox{-10ex}[0pt]{\begin{sideways}Summary $y$\end{sideways}} & \cellcolor{Gray}
  \textcolor{blue}{
  	%It takes 5-10 minutes to oceans drive with cab or car. 
  	At first they
    took us to an unready room which was disappointing but after a short
    wait they took us to a really big room with a great harbor scene as an
    apology to the mess. The rooms are pretty new or renovated
    recently. Bathroom is clean and wide. The beds are comfortable and
    big. 
% The breakfast buffet wasn't included in the price. 
%but we enjoyed
%    continental breakfast.
} \\ 
  \multicolumn{2}{c}{} \\
  \raisebox{-9ex}[0pt]{\begin{sideways}Review $x_1$\end{sideways}} &  \cellcolor{LightCyan}
  Check in was quick and our bags were brought to the
  room in a timely manner. \bleuline{The \textcolor{blue}{\keyword{rooms}}
    and \textcolor{blue}{\keyword{hallways}} left a little more to be
    desired.} \bleuline{The \textcolor{blue}{\keyword{rooms}} didnt look
    nearly as good as they did in other less known cities.} \bleuline{No
    safe or \textcolor{blue}{\keyword{frig}} in the
    \textcolor{blue}{\keyword{rooms.}}} The staff was great. \\
\multicolumn{2}{c}{} \\
  \raisebox{-9ex}[0pt]{\begin{sideways}Review $x_2$\end{sideways}} & \cellcolor{LightCyan} 
    Only option for a hot meal for
    breakfast was scrambled eggs and bacon; The toaster was broken as
    well, with burned out elements. Other food in the lounge was good
    (fruit, coffee). Recommendation: eat elsewhere; even
    \textcolor{blue}{\keyword{room service}} would probably have been
    better. 
\end{tabular}
\end{small}
\vspace*{-1ex}
\caption{Pseudo-summary~$y$ and input reviews~$X$; the aspect code for
  summary~$y$ is \aspect{\textcolor{blue}{room}}. Review sentences
  with the same aspect are \bleuline{underlined} and same aspect-keywords
  are  \textcolor{blue}{\keyword{magnified}}.} 
%          induced. The aspect code is first induced from the
%          summary. Then, aspect keywords and sentences are induced
%          from the reviews, which are then ranked and selected using
%          the given aspect code. At test time, we generate different
%          kinds of summaries by modifying aspect codes based on user
%          request.  }
	\label{fig:controller-induction}
\end{figure}

\subsection{Synthetic Dataset Creation}
\label{sec:dataset-creation}

The MIL model allows us to learn three kinds of aspect controllers
which are subsequently used to create a synthetic dataset for training
our summarizer. These are \emph{aspect codes}, essentially
document-level aspect predictions~$\mathbf{z}_{D}$, which
control the overall aspect of the summary, \emph{aspect keywords}
ensure content support by explicitly highlighting which tokens from
the input should appear in the summary, and \emph{aspect-relevant
  sentences} which provide textual context for summary generation
(while non-aspect-related sentences are ignored). %We will now explain
%how the synthetic dataset is created.

We first sample review~$r_i$ as a pseudo-summary from review set~$R_e$
of entity~$e$. We treat~$r_i$ as a pseudo-summary provided it
resembles a real summary. We assume that opinion summaries discuss
specific aspects regarding entity~$e$. We use our controller induction
model to verify this, i.e.,~document-level aspect predictions
$\mathbf{z}_{\mathcal{D}}$ for $r_i$ should be positive for at least
one aspect. Provided $r_i$ fulfills this constraint, we use it as
summary $y$ and $R_e - \{r_i\}$ as review set~$X$. A simplified
example is shown in Figure~\ref{fig:controller-induction}, the pseudo
summary is highlighted in gray and the input reviews in cyan. The
summary focuses on the \aspect{room} aspect of a hotel and this is its
aspect code (shown in blue).

Let $(X,y)$ denote review set~$X$ for summary~$y$ (we only show two
reviews in Figure~\ref{fig:controller-induction} but there are usually
hundreds). We obtain (positive) document-level aspect
predictions~$\mathbf{z}_{D}^{(y)}$ for summary~$y$ and sentence-level
aspect predictions $\mathbf{z}_{\mathcal{S}}^{(x)}$ for all reviews $x
\in X$. We then rank review sentences in~$X$ based on their similarity
to the summary's overall aspect. Specifically, we compare predictions
$\mathbf{z}_{\mathcal{S}}^{(x)}$ with $\mathbf{z}_{\mathcal{D}}^{(y)}$
using the soft margin loss function from
Equation~\eqref{eq:softmargin}. We also compare token-level
predictions $\mathbf{z}_{\mathcal{T}}^{(x)}$ with
$\mathbf{z}_{\mathcal{D}}^{(y)}$ using the same function to induce aspect
keywords. In
Figure~\ref{fig:controller-induction} sentences which discuss the same
aspect as the summary are underlined, and same-aspect keywords are
magnified. For illustration purposes we only show one aspect code in
Figure~\ref{fig:controller-induction}, but these can be several, and
different review sentences and keywords would be selected for
different aspects.

 %\textbf{The second step is to induce
 %the aspect controllers $z$ using the method described in the previous
 %paragraph.}

\subsection{Opinion Summarization Model}
\label{sec:opinion-summarization}

We use a pretrained sequence-to-sequence Transformer
model \cite{raffel2020exploring} to generate opinion summaries. We
transform the aspect controllers~$z$ into the following format:

\begin{quote}
\texttt{[CODE]} \texttt{[ASPECT$_{2}$]} \texttt{[ASPECT$_{3}$]}
\texttt{[KEY]} keyword$_{1}$ keyword$_{2}$ ... \texttt{[SNT]} first sentence \texttt{[SNT]} second sentence ... 
\end{quote}
where \texttt{[CODE]}, \texttt{[KEY]}, and \texttt{[SNT]} are
indicators denoting that the next tokens correspond to aspect codes,
keywords, and review sentences.

Instead of the full set of input reviews $X$, the encoder takes~$z$ as input and produces multi-layer
encodings~$\mathbf{Z}$. The decoder then outputs a token distribution
$p(y_t)$ for each time step $t$, conditioned on both $\mathbf{Z}$ and
$y_{1:t-1}$ through attention:
\begin{align}
	\label{eq:generate}
	\mathbf{Z} &= \text{Encoder}(z) \\
	p(y_t) &= \text{Decoder}(y_{1:t-1}, \mathbf{Z})
\end{align}

We fine-tune the model using a maximum likelihood loss to optimize the
probability distribution~$p(y)$ based on gold summary $\hat{y}$:
\begin{equation}
	\mathcal{L}_{gen} = -\sum\nolimits_t \hat{y}_t \log p(y_t)
\end{equation}

During inference, we can generate different kinds of opinion summaries
by modifying the aspect controllers. When creating a general summary,
we use all aspect codes as input. Analogously, when generating a
single aspect summary, we use one aspect code. The aspect codes guide
the selection of keywords and sentences from the input reviews (see
Figure~\ref{fig:controller-induction}) which are given as input to our
Transformer model to generate the summary (see
Equation~\eqref{eq:generate}).

\section{Experimental Setup}

\paragraph{Datasets}

\begin{table}[t]
	\centering
	\small
	\begin{tabular}{@{}lrr@{}}
		\thickhline
		Dataset & \textsc{Space} & \textsc{Oposum+} \\
		\thickhline
		review corpus size & 1.14M & 4.13M \\
		\#domains & 1~~~ & 6~~~ \\
		\#aspects & 6~~~ & 18~~~ \\
		\#test examples & 50~~~ & 60~~~ \\
		\#reviews/example & 100~~~ & 10~~~ \\
		\#summaries/example & 3~~~ & 3~~~ \\
		\#general summaries & 150~~~  & \uline{180}~~~  \\
		\#aspect summaries & 900~~~  & 540~~~  \\
		\thickhline
	\end{tabular}
\vspace*{-1ex}
	\caption{Statistics for \textsc{Space} and \textsc{Oposum++}
	(\uline{underlined} summaries are extractive). }
	\label{tab:data}
\end{table}

We performed experiments on two opinion summarization datasets
covering different review domains. \textsc{Space}
\cite{angelidis2020extractive} is a large corpus of ``hotel'' reviews
from TripAdvisor; it contains human-written abstractive opinion
summaries for evaluation only.  Each instance in the evaluation set
consists of 100 reviews and seven summaries: one general summary and
six aspect-specific ones representing the aspects \aspect{building},
\aspect{cleanliness}, \aspect{food}, \aspect{location},
\aspect{rooms}, and \aspect{service}.  \textsc{Oposum}
\cite{angelidis2018summarizing} is a large corpus of product reviews
from six different domains: ``laptop bags'', ``bluetooth headsets'',
``boots'', ``keyboards'', ``televisions'', and ``vacuums''. It also
includes an evaluation set with extractive general summaries. We
extended this dataset by (a) adding aspect-specific summaries which
are human-written and abstractive following the methodology from
\citet{angelidis2020extractive}, and (b) increasing the size of the
corpus.  We call this extended dataset \textsc{Oposum+}.  Both
datasets include five human-annotated seed words for each aspect (see
Appendix for details). Data statistics are shown in
Table~\ref{tab:data}.  Using our synthetic dataset creation method, we
were able to generate 512K and 341K training instances for
\textsc{Space} and \textsc{Oposum+}, respectively.

\paragraph{Implementation}

For our pretrained Transformer models, we used weights and settings
available in the HuggingFace library
\cite{wolf2020transformers}. Specifically, we used
\texttt{distilroberta-base} \cite{liu2019roberta,sanh2019distilbert}
as our language model and \texttt{t5-small} \cite{raffel2020exploring}
as our sequence-to-sequence model. We trained the controller induction
model with a learning rate of~$1e-4$ for 100K steps, using $h=12$
heads. For \textsc{Oposum+}, we trained separate controller induction
models for different domains.  %Additionally, we found that the model
%tends to focus its prediction towards a single token, disregarding
%other potentially correct tokens. %To solve this, we warm up the first
%10K steps of training, masking~20\% of the tokens.  
For the aspect controllers, we selected 10-best keywords, and review
sentences were truncated up to 500~tokens to fit in the pretrained
model.  For summarization, we used a learning rate of~$1e-6$ and~500K
training steps.  We used Adam with weight decay
\cite{loshchilov2018decoupled} to optimize both models. We added a
linear learning rate warm-up for the first 10K steps.  We generate
summaries with beam search of size 2 and refrain from repeating ngrams
of size~3 \cite{paulus2018a}.

\section{Results}

We compared our \textbf{A}spect \textbf{C}ontroll\textbf{e}d
\textbf{Sum}marization (\textsc{AceSum}) model with several extractive
and abstractive approaches.  Traditional extractive systems include
selecting as a summary the review closest to the \textsc{Centroid}
\cite{radev2004centroid} of the input reviews and \textsc{LexRank}
\cite{erkan2004lexrank}, a PageRank-like algorithm that selects the
most salient sentences from the input.  For both methods we used BERT
encodings \cite{devlin2019bert} to represent sentences and documents.
Other extractive systems include \textsc{QT}\footnote{We report
  results for QT using our seed words which are human-annotated.  We
  also present results in the Appendix with their seed words which
  were automatically induced.}  \cite{angelidis2020extractive}, a
neural clustering method that uses Vector-Quantized Variational
Autoencoders \cite{vandenoord2017neural} to represent opinions in
quantized space, and \textsc{AceSumExt}, an extractive version of our
model that uses sentences ranked by our controller induction model as
input (truncated up to 500~tokens) to LexRank.

Abstractive systems include \textsc{MeanSum} \cite{chu2019meansum}, an
autoencoder that generates summaries by reconstructing the mean of
review encodings, \textsc{Copycat} \cite{bravzinskas2019unsupervised},
a hierarchical variational autoencoder which learns a latent code of
the summary, and two variants of T5 \cite{raffel2020exploring} trained
with different synthetic dataset creation methods. For
\textsc{T5-random}, summaries are randomly sampled
\cite{bravzinskas2019unsupervised}, whereas for \textsc{T5-similar}
reviews are sampled based on their similarity to a candidate summary
\cite{amplayo2020unsupervised}.

Finally, we compared against two upper bounds: an extractive
\textsc{Oracle} which selects as a summary the review with the best
ROUGE score against the input, and a \textsc{Human} upper bound,
calculated as inter-annotator ROUGE. Examples generated by our model
are in Table~\ref{tab:examples} and the Appendix.

\subsection{Automatic Evaluation}
\label{sec:auto-results}

We evaluated the quality of general and aspect-specific opinion
summaries using $\text{F}_1$ ROUGE
\cite{lin-hovy-2003-automatic}. Unigram and bigram overlap (ROUGE-1/2)
are proxies for assessing informativeness while the longest common
subsequence (ROUGE-L) measures fluency.

\paragraph{General Opinion Summarization}

Table~\ref{tab:generic_results} reports results on general opinion
summarization. As can be seen, \textsc{AceSum} outperforms all
competing models on \textsc{Space} and performs best among abstractive
systems on \textsc{Oposum+}. Our extractive model, \textsc{AceSumExt},
is overall best on \textsc{Oposum+}. This is expected since general
\textsc{Oposum+} summaries are extractive. Amongst abstractive models,
Transformer-based models outperform \textsc{MeanSum} and
\textsc{Copycat}, demonstrating that pretraining is helpful for
opinion summarization.

\begin{table}[t]
	\small
	\centering
	\begin{tabular}{@{}l@{}c@{}c@{}c@{}c@{}c@{}c@{}}
		\thickhline
		& \multicolumn{3}{@{}c@{}}{\textsc{Space}} & \multicolumn{3}{@{}c@{}}{\textsc{Oposum+}}\\
		\multicolumn{1}{c}{Model} &  R1    & R2    & RL & R1 & R2 & RL \\
		\thickhline
		\textsc{Centroid} & 31.29  & 4.91  & 16.43  & 33.44  & 11.00  & 20.54 \\
		\textsc{LexRank} & 31.41 & 5.05 & 18.12 & 35.42 & 10.22 & 20.92  \\
		\textsc{QT} & 38.66 & 10.22 & 21.90 &37.72 & 14.65 & 21.69 \\
		\textsc{AceSumExt} & 35.50 & 7.82 & 20.09 &  \hspace*{.15cm}\textbf{38.48}${}^*$ & \hspace*{.15cm}\textbf{15.17}${}^*$ & \hspace*{.15cm}\textbf{22.82}${}^*$ \\
		\hline\hline
		\textsc{MeanSum}  & 34.95 & 7.49 & 19.92  &  26.25  & 4.62 & 16.49 \\
		\textsc{Copycat}  & 36.66  & 8.87  & 20.90  & 27.98 & 5.79 & 17.07    \\
		\textsc{T5-random}  & 37.65 & 10.62 & 22.82 & 29.88 & 5.64 & 17.19 \\
		\textsc{T5-similar}  & 38.84 & 10.82 & 22.74 & 30.42 & 6.07 & 17.17  \\
		\textsc{AceSum} & \hspace*{.15cm}\textbf{40.37}${}^*$ & \hspace*{.15cm}\textbf{11.51}${}^*$ & \hspace*{.15cm}\textbf{23.23}\hspace*{.15cm} & 32.98 & 10.72 & 20.27 \\
		\hline \hline
		\textsc{Oracle} & 40.23 & 13.96 & 23.46 & 41.88 & 21.52 & 29.30 \\
		\textsc{Human} & 49.80  & 18.80  & 29.19  & 55.42  & 37.26  & 44.85  \\
		\thickhline
	\end{tabular}%
\vspace*{-1ex}
\caption{Automatic evaluation for \emph{general summarization}.
  Extractive/Abstractive/Upper-bound models are shown in first/second/third
  block. Best systems are boldfaced; an asterisk (*)~means there is a significant difference
  between best and 2nd best systems (based on paired bootstrap
  resampling; $p < 0.05$).}
	\label{tab:generic_results}%
\end{table}%

\paragraph{Aspect-Specific Opinion Summarization}

Most comparison systems (all except \textsc{QT}) cannot naturally
generate aspect-specific summaries.  We use a simple
sentence-filtering method to remove non-aspect-related sentences from
the input during inference. Specifically, we use BERT encodings
\cite{devlin2019bert} to represent tokens in review sentences
$\{r^{(bert)}_i\}$ and aspect seeds~$\{a^{(bert)}_j\}$. We then rank
the review sentences based on the maximum similarity between seed and
sentence tokens, calculated as $\max_{i,j}
(\operatorname{sim}(r^{(bert)}_i, a^{(bert)}_j))$, where
$\operatorname{sim}(a,b)$ is the cosine similarity function.  This
method cannot be ported to the \textsc{Centroid} and \textsc{Oracle}
baselines, and thus we do not compare with them.

Our results are summarized in Table~\ref{tab:aspect_results}. Note
that \textsc{Space} and \textsc{Oposum+} focus exclusively on
\emph{single} aspect summaries. We assess our model's ability to
generate summaries covering multiple aspects in the following section.
Overall, \textsc{AceSum} performs best across datasets and metrics,
which shows that our controllers can effectively customize summaries
based on aspect queries. Interestingly, amongst extractive models,
\textsc{AceSumExt} performs best. This suggests that, a simple
centrality-based extractive approach such as LexRank
\cite{erkan2004lexrank} can produce good enough summaries as long as
an effective sentence filtering method is applied beforehand (in our
case this is based on the controller induction model). T5 models
perform substantially worse on this task, indicating that synthetic
datasets based on either random or similarity-based sampling
techniques are not suited to aspect-specific opinion summarization.

\begin{table}[t]
	\small
	\centering
	\begin{tabular}{@{}l@{}c@{}c@{}c@{}c@{}c@{}c@{}}
		\thickhline
		& \multicolumn{3}{@{}c@{}}{\textsc{Space}} & \multicolumn{3}{@{}c@{}}{\textsc{Oposum+}}\\
		\multicolumn{1}{c}{Model} &  R1    & R2    & RL & R1 & R2 & RL \\
		\thickhline
		\textsc{LexRank} & 24.61 & 3.41 & 18.03 & 22.51 & 3.35 & 17.27  \\
		\textsc{QT} & 28.95 & 8.34 & 21.77 & 23.99 & 4.36 & 16.61 \\
		\textsc{AceSumExt} & 30.91 & 8.77 & 23.61 & 26.16 & 5.75 & 18.55 \\
		\hline\hline
		\textsc{MeanSum}  & 25.68 & 4.61 & 18.44 & 24.63 & 3.47 & 17.53 \\
		\textsc{Copycat}  & 27.19 & 5.63 & 19.18 & 26.17 & 4.30 & 18.20   \\
		\textsc{T5-random}  & 21.40 & 4.83 & 15.45 & 24.47 & 4.20 & 16.18 \\
		\textsc{T5-similar}  & 22.69 &  5.12 & 16.44 & 23.86 & 4.30 & 16.36 \\
		\textsc{AceSum} & \hspace*{.15cm}\textbf{32.41}${}^*$ & \hspace*{.15cm}\textbf{9.47}${}^*$ & \hspace*{.15cm}\textbf{25.46}${}^*$ & \hspace*{.15cm}\textbf{29.53}${}^*$ & \hspace*{.15cm}\textbf{6.79}${}^*$ & \hspace*{.15cm}\textbf{21.06}${}^*$ \\
		\hline \hline
		\textsc{Human} & 44.86  & 18.45  & 34.58  &  43.03  &  16.16  & 31.53  \\
		\thickhline
	\end{tabular}%
\vspace*{-1ex}
\caption{Automatic evaluation for \emph{aspect-specific
    summarization}.  Extractive/Abstractive/Upper-bound models
  are shown in first/second/third
  block. Best systems are  boldfaced; an asterisk (*)~means there is a significant difference
  between best and 2nd best systems (based on paired bootstrap
  resampling; $p < 0.05$).}
	\label{tab:aspect_results}%
\end{table}%

\paragraph{Ablation Studies}

We present various ablation studies on the controller induction model
and the summarization model itself. In Table~\ref{tab:ablate_mil}, we
compare our multiple instance pooling (MIP) mechanism with three
standard pooling methods: mean, max, and attention-based pooling.  We
evaluate models using document and sentence F${}_1$ which measures the
quality of document- and sentence-level aspect predictions. We
extrapolate aspect labels for documents and sentences from the
development set which contains aspect-specific summaries. We assume
the aspect for which a summary is written is the document label and
that all sentences within the summary are also representative of the
same aspect.  Results show that attention and mean pooling are not
suitable for multi-instance learning, underperforming especially on
document-level F${}_1$. This suggests that token-level predictions are
not used effectively to predict higher level aspects. Our results
confirm that using multiple experts (i.e.,~attention heads) yields
better aspect predictions.

\begin{table}[t]
	\small
	\centering
	\begin{tabular}{@{}lcccc@{}}
		\thickhline
		 & \multicolumn{2}{c}{\textsc{Space}} & \multicolumn{2}{c}{\textsc{Oposum+}} \\
		\multicolumn{1}{c}{Model} & Doc F${}_1$ & Sent F${}_1$ & Doc F${}_1$ & Sent F${}_1$ \\
		\thickhline
		MIP (ours) & \textbf{77.35} & \textbf{40.85} & \textbf{83.28} & \textbf{50.48} \\
		Max & 63.35 & 35.12 & 66.52 & 44.00 \\
		Attention & 31.77 & 29.30 & 34.00 & 35.80 \\
		Mean & 27.38 & 27.87 & 30.38 & 34.35 \\
		\thickhline
	\end{tabular}
\vspace*{-1ex}
\caption{Performance of controller
  induction models (document- and sentence-level); comparison of multiple instance pooling (MIP)
  against max, mean, and attention pooling.} 
	\label{tab:ablate_mil}%
\end{table}

In Table~\ref{tab:ablate_sum}, we evaluate the contribution of
different aspect controllers to summarization output.  Selecting
sentences randomly rather than based on aspect hurts performance, in
particular when generating aspect-specific summaries.  We also find
that aspect codes substantially increase model performance in
\textsc{Oposum+}. We conjecture that this is due to \textsc{Oposum+}
having multiple domains and, consequently, more aspects compared to
\textsc{Space}.

\begin{table}[t]
	\small
	\centering
	\begin{tabular}{@{}l@{~}c@{~~~}c@{~~~}c@{~~~}c@{}}
		\thickhline
		& \multicolumn{2}{c}{\textsc{Space}} &  \multicolumn{2}{c}{\textsc{Oposum+}}\\
	\multicolumn{1}{c}{Model} 	& General & Aspect & General & Aspect \\
		\thickhline
		\textsc{AceSum} & 23.23 & 25.03 & 19.64 & 20.16 \\
		\quad No aspect code & 22.29 & 24.99 & 17.22 & 17.54\\
		\quad No keywords & 21.88 & 24.82 & 18.97 & 19.97 \\
		\quad Random sentences & 22.42 & 19.16 & 18.96 & 13.44 \\
		\thickhline
	\end{tabular}
\vspace*{-1ex}
\caption{Variants of \textsc{AceSum} with different aspect
  controllers. Results are shown using ROUGE-L for general and aspect-specific opinion  summaries.} 
	\label{tab:ablate_sum}%
\end{table}

\subsection{Human Evaluation}

We conducted several human elicitation studies 
to further analyze the summaries produced by competing systems
using the Amazon
Mechanical Turk crowdsourcing platform. 

\paragraph{Best-Worst Scaling}
The first study assessed the
quality of general opinion summaries using Best-Worst Scaling (BWS;
\citealp{louviere2015best}). Participants were shown a human-written
summary, in relation to which they were asked to select the best and
worst among system summaries, taking into account the following
criteria: \textit{Informativeness} (how consistent are the opinions
with the reference?), \textit{Coherence} (is the summary easy to read
and well-organized?), \textit{Conciseness} (does the summary provide
useful information in a concise manner?), and \textit{Fluency} (is the
summary grammatical?).

We compared general summaries produced by the two best performing
extractive (\textsc{LexRank}, \textsc{QT}) and abstractive
(\textsc{T5-similar}, \textsc{AceSum}) systems according to ROUGE.  We
elicited three judgements for all entities in the \textsc{Space} and
\textsc{Oposum+} test sets. Table~\ref{tab:human-eval-generic}
summarizes our results.  BWS values range from $-100$ (unanimously
worst) to $100$ (unanimously best). \textsc{AceSum} is deemed best for
all criteria on both datasets. Crowdworkers also rated \textsc{QT}
high on informativeness, which indicates that aspect modeling is
helpful, but low on other criteria (e.g.,~coherence and conciseness)
due to its extractive nature.

\begin{table}[t]
	\small
	\centering
	\begin{tabular}{@{}ld{2.2}d{2.2}d{2.2}d{2.2}@{}}
		\thickhline
%		& \multicolumn{4}{@{}c@{}}{\textsc{Space}} \\
		\multicolumn{1}{@{}c}{\textsc{Space}} & \mc{1}{c}{\textrm{Inf}} & \mc{1}{c}{\textrm{Coh}} & \mc{1}{c}{\textrm{Con}} & \mc{1}{c}{\textrm{Flu}} \\
		\thickhline
		\textsc{LexRank} & -48.3 & -38.4 & -36.9 & -43.3 \\
		\textsc{T5-similar} & 5.8 & 11.2 & 17.2 & 0.6 \\
		\textsc{QT} & 20.4 & 1.3 & 1.2 & 2.6 \\
		\textsc{AceSum} & 22.1 & 26.0^* & 18.5 & 38.8^* \\
		\thickhline
\multicolumn{5}{c}{} \\ \thickhline
		\multicolumn{1}{@{}c}{\textsc{Oposum+}} & \mc{1}{c}{\textrm{Inf}} & \mc{1}{c}{\textrm{Coh}} & \mc{1}{c}{\textrm{Con}} & \mc{1}{c}{\textrm{Flu}} \\
		\thickhline
		\textsc{LexRank} &  -27.3 & -21.1 & -18.2 & -23.8\\
		\textsc{T5-similar} & -31.1 & 10.0 & 4.7 & -1.9\\
		\textsc{QT} & 20.3 & -25.3 & -21.6 & -9.6  \\
		\textsc{AceSum} & 38.1^* & 36.3^* &
                35.2^* & 35.3^*\\
		\thickhline
	\end{tabular}
\vspace*{-1ex}
	\caption{\emph{Best-Worst Scaling} evaluation. Best values are
	bold-faced. An asterisk (*) means that the system is
	significantly better than the second best system (one-way
	ANOVA with posthoc Tukey HSD tests, $p<0.05$). Inf:
	informative, Coh: coherent, Con: concise, Flu: fluent.} 
	\label{tab:human-eval-generic}%
\end{table}

\paragraph{Aspect Controllability}

We also conducted a user study to assess the quality of
aspect-specific summaries. We showed participants the aspect in
question as well as aspect summaries from \textsc{T5-similar},
\textsc{QT}, \textsc{AceSum}, and \textsc{Human}. Crowdworkers were
asked to decide whether the summaries discussed the given aspect
\textit{exclusively}, \textit{partially}, or \textit{not at all}.  We
elicited three judgments for all test entities.  As can be seen in
Table~\ref{tab:human-eval-aspect}, \textsc{Space} summaries produced
by \textsc{AceSum} exclusively discuss a single aspect 50.9\% of the
time.  \textsc{T5-similar} mostly produces general summaries (74.8\%
of them partially discuss the given aspect) which is not surprising,
given that it has no special-purpose mechanism for modeling aspect.
QT summaries are more topical for the opposite reason.  In general,
automatic systems perform worse on \textsc{Oposum+} whose larger
number of domains renders this dataset more challenging. Finally, we
observe a big gap between model and \textsc{Human} performance.

\begin{table}[t]
	\small
	\centering
	\begin{tabular}{@{}lrrr@{}}
		\thickhline
%		& \multicolumn{3}{@{}c@{}}{\textsc{Space}} \\
		\multicolumn{1}{@{}c}{\textsc{Space}} & Exclusive & Partial & None \\
		\thickhline
		\textsc{T5-similar} & 10.6 & 74.8 & 14.6 \\
		\textsc{QT} & 43.8 & 39.0 & 17.1 \\
		\textsc{AceSum} & 50.9 & 42.6 & 6.5 \\
		\textsc{Human} & 64.9 & 31.6 & 3.5\\
		\thickhline
		& \multicolumn{3}{@{}c@{}}{\textsc{}} \\\thickhline
		\multicolumn{1}{@{}c}{\textsc{Oposum+}} & Exclusive &
		Partial & None \\ \thickhline \textsc{T5-similar} &
		9.4 & 48.2 & 42.5 \\ \textsc{QT} & 22.2 & 41.9 &
		35.9 \\ \textsc{AceSum} & 42.2 & 45.4 &
		12.4 \\ \textsc{Human} & 63.0 & 31.5 &
		5.6 \\ \thickhline \end{tabular} 
\vspace*{-1ex}
\caption{Proportion
		of  summaries that discuss the target aspect
		exclusively, partially, or not
		at all.}  \label{tab:human-eval-aspect}%
\end{table}

We further verified whether \textsc{AceSum} can produce summaries
covering two aspects. Although it can generate summaries with more
aspects (see Table~\ref{tab:examples}), we hypothesize that user
queries pertaining to two aspects would be most frequent. Besides, if
performance with two aspects is inferior, there is little chance it
will improve with more aspects. For each test example we elicited
three judgments and randomly selected two aspect pairs from the set of
all possible aspect combinations.  We compared \textsc{AceSum} against
\textsc{QT} (for which we used seed words representing both target
aspects).  Participants were shown the two aspects and the
summaries generated by \textsc{QT} and \textsc{AceSum}. They were
asked to decide whether the summaries discussed (a)~both target
aspects exclusively (b)~one of the aspects (c)~other aspects in
addition to the target ones, and (d)~none of the two aspects. The
results in Table~\ref{tab:human-eval-multi} show that \textsc{AceSum}
is able to produce two-aspect summaries effectively 61.3\% of the time
on \textsc{Space} and 47.0\% of the time on
\textsc{Oposum+}. \textsc{QT} on the other hand mostly creates
single-aspect summaries.

\begin{table}[t]
\small
\centering
\begin{tabular}{@{}lrrrr@{}}
	\thickhline
	\textsc{Space} & All & One & Other & None \\
	\thickhline
	\textsc{QT} & 10.0 & 35.3 & 34.7 & 20.0\\
	\textsc{AceSum} & 61.3 & 19.3 & 18.0 & 1.3 \\
	\thickhline
\mc{1}{c}{} \\ \thickhline
	\textsc{Oposum+} & All & One & Other & None \\
	\thickhline
	\textsc{QT} & 18.8 & 27.5 & 33.6 & 20.1 \\
	\textsc{AceSum} & 47.0 & 16.8 & 26.8 & 9.4 \\
	\thickhline
\end{tabular}
\vspace*{-1ex}
\caption{Proportion of target aspects discussed in system summaries (All:
  both  aspects are mentioned; One: only one  is
  mentioned; Other: other aspects are also mentioned; None: no  aspects are mentioned).}
\label{tab:human-eval-multi}%
\end{table}

%\subsection{System Outputs}

\paragraph{Summary Veridicality}
Our third study examined the veridicality of the generated summaries,
i.e.,~whether the opinions mentioned in them are indeed discussed in
the input reviews. Participants were shown reviews and corresponding
system summaries and were asked to verify, for each sentence of the
summary, whether it was fully supported by the reviews, partially
supported, or not at all supported. We performed this experiment on
\textsc{Oposum+} only since the number of reviews is small and
participants could read them all in a timely fashion.  We collected
three judgments for all system summaries, both general and
aspect-specific ones.  Participants assessed the summaries produced by
\textsc{T5-similar} and \textsc{AceSum}. We also included
\textsc{Gold}-standard summaries as an upper bound but no output from
an extractive system as it by default produces veridical summaries
which contain facts mentioned in the reviews.

Table~\ref{tab:veridicality} reports the percentage of fully
(FullSupp), partially (PartSupp), and un-supported (NoSupp) sentences.
Not unsurprisingly, \textsc{Gold} summaries display the highest
percentage of fully supported sentences for both general and
aspect-specific summaries.  \textsc{AceSum} and \textsc{T5-similar}
present similar proportions of supported sentences when it comes to
general summaries, with \textsc{AceSum} having a slight advantage.
The proportion of supported sentences is higher in aspect summaries
for \textsc{T5-similar}. Note that this model struggles to actually
generate aspect-specific summaries (see
Table~\ref{tab:human-eval-aspect}); instead, it generates any-aspect
summaries which maybe veridical but off-topic.

%slightly
%better supported than \textsc{T5-similar} when comparing general
%summaries. We note that \textsc{T5-similar} cannot produce
%aspect-specific summaries, thus comparisons between \textsc{AceSum}
%and \textsc{T5-similar} on aspect-specific summaries may be
%unreliable.

\begin{table}[t]
	\small
	\centering
	\begin{tabular}{@{}lrrr@{}}
		\thickhline
		\multicolumn{4}{@{}c@{}}{\textsc{Oposum+} General} \\
		\multicolumn{1}{@{}c}{Model} & FullSupp & PartSupp & NoSupp \\
		\thickhline
		\textsc{T5-similar} & 53.3 & 36.9 & 9.8 \\
		\textsc{AceSum} & 59.9 & 32.2 & 8.0 \\
		\textsc{Human} & 88.4 & 7.0 & 4.6\\
		\thickhline
		\\\thickhline
		\multicolumn{4}{@{}c@{}}{\textsc{Oposum+} Aspect} \\
		\multicolumn{1}{@{}c}{Model} & FullSupp & PartSupp & NoSupp \\ \thickhline 
		\textsc{T5-similar} & 57.3 & 29.4 & 13.3 \\ 
		\textsc{AceSum} & 54.2 & 32.3 & 13.5 \\ 
		\textsc{Human} & 67.8 & 20.7 & 11.6 \\ 
		\thickhline 
	\end{tabular} 
	\vspace*{-1ex}
	\caption{\emph{Summary veridicality} evaluation. Proportion
          of summaries that are
          fully supported, partially supported, or not
          supported at all.}  \label{tab:veridicality}%
\end{table}

\section{Conclusions}
%: given a set of reviews and a
%query focusing on one or multiple aspects, we are tasked to generate
%an opinion summary that discusses the target aspects.

In this work, we presented an abstractive approach to
aspect-controlled opinion summarization.  Key to our model is the
induction of aspect controllers which facilitate the creation of a
synthetic training dataset and guide summary generation towards the
designated aspects.  Extensive experiments on two benchmarks show that
our model achieves state of the art across the board, for both general
and aspect-specific opinion summarization. %We also show that our model
%can effectively generate multi-aspect summaries, simply by modifying
%the aspect controllers.

In the future, we would like to focus on controlling additional facets
of opinion summaries such as sentiment or length. It would also be
interesting to learn aspects %(and their corresponding seed words)
from data rather than specifying them apriori as well as dealing with
unseen aspects (e.g.,~in a scenario where reviews discuss new features
of a product). 

%more effective methods for aspect modeling to improve summary
%generation in settings with many aspects and to possibly allow
%aspect-specific summarization for unseen aspects.  Finally, we believe
%that our work is a necessary step towards future work on personalized
%opinion mining. It would be interesting to apply the techniques in
%this paper to generate more user-tailored opinion summaries.

\paragraph{Acknowledgments}

We thank the anonymous reviewers for their feedback. We gratefully
acknowledge the support of the European Research Council (Lapata,
award number 681760). The first author is supported by a Google PhD
Fellowship.

\bibliographystyle{acl_natbib}
\bibliography{emnlp2021}

\clearpage
\appendix
\section{Appendix}

\subsection{List of Seed Words}

Tables \ref{tab:seed-words1} and \ref{tab:seed-words2} shows the seed words we used in our
experiments.  These were generated semi-automatically: we first
obtained aspect-specific words through the automatic method introduced
in \citet{angelidis2018summarizing} and
\citet{angelidis2020extractive} and then asked human annotators to
filter out the noise (i.e.,~words that were assigned incorrect
aspects).

\begin{table}[t]
	\small
	\centering
	\begin{tabular}{@{}lc@{}}
		\thickhline
		Aspect & ``Hotels'' \\
		\hline
		\aspect{building} & lobby pool decor gym area \\
		\aspect{cleanliness} & clean spotless garbage dirty stain \\
		\aspect{food} & breakfast food buffet restaurant meal \\
		\aspect{location} & location walk station distance bus \\
		\aspect{rooms} & room bed bathroom shower spacious \\
		\aspect{service} & staff service friendly helpful desk \\
		\thickhline
\end{tabular}
	\caption{\textsc{Space} seed words for the ``Hotels''
          domain.
	\label{tab:seed-words1}
}
\end{table}

\begin{table}[t]
	\small
	\centering
	\begin{tabular}{@{}lc@{}}
		\thickhline
		Aspect & ``Laptop Bags'' \\		\hline
		\aspect{looks} & looks color stylish looked pretty \\
		\aspect{quality} & quality material poor broke durable \\
		\aspect{size} & fit fits size big space \\
		\thickhline
\multicolumn{2}{c}{} \\ \thickhline
		Aspect & ``Bluetooth Headsets''\\	\hline
		\aspect{comfort} & ear fit comfortable fits buds \\
		\aspect{ease of use} & easy button simple setup control \\
		\aspect{sound quality} & sound quality hear noise volume \\
		\thickhline
\multicolumn{2}{c}{} \\ \thickhline
		Aspect & ``Boots'' \\	\hline
		\aspect{comfort} & comfortable foot hurt ankle comfy \\
		\aspect{looks} & cute look looked fringe style \\
		\aspect{size} & size half big little bigger \\
		\thickhline
\multicolumn{2}{c}{} \\ \thickhline
		Aspect & ``Keyboards'' \\ 		\hline
		\aspect{build quality} & working months build stopped quality \\
		\aspect{feel/comfort} & feel comfortable feels mushy shallow \\
		\aspect{layout} & key keys delete backspace size \\
		\thickhline
\multicolumn{2}{c}{} \\ \thickhline
		Aspect & ``Televisions'' \\\hline
		\aspect{connectivity} &hdmi computer port usb internet \\
		\aspect{image quality} & picture color colors bright clear \\
		\aspect{sound quality} & sound speakers loud tinny bass \\
		\thickhline
\multicolumn{2}{c}{} \\ \thickhline
		Aspect & ``Vacuums''\\ \hline
		\aspect{accessories} & filter brush attachments attachment turbo \\
		\aspect{ease of use} & easy push corners awkward impossible \\
		\aspect{suction power} & suction powerful power hair quiet \\
		\thickhline
	\end{tabular}
	\caption{\textsc{Oposum+} seed words for various domains and
          their aspects.}
	\label{tab:seed-words2}%
\end{table}

\subsection{Results using Automatic Seed Words}

Table \ref{tab:auto-seeds-results} shows comparisons between
\textsc{AceSum} and \textsc{QT} using automatically generated seed
words for aspect-specific summarization (as used in \citealp{angelidis2020extractive}).  Our model performs better
than \textsc{QT} on both datasets, while both models benefit from
better quality seed words with noticeable increase in ROUGE scores.

\begin{table}[t]
	\small
	\centering
	\begin{tabular}{@{}lcccccc@{}}
		\thickhline
		& \multicolumn{3}{@{}c@{}}{\textsc{Space}} & \multicolumn{3}{@{}c@{}}{\textsc{Oposum+}}\\
		\multicolumn{1}{c}{Model} &  R1    & R2    & RL & R1 & R2 & RL \\
		\thickhline
		\multicolumn{7}{@{}c@{}}{\textit{using automatic seed words}} \\
		\hline
		\textsc{QT} & 28.95 & 8.34 & 21.77 & 23.16 & 4.13 & 16.81 \\
		\textsc{AceSum} & 30.78 & 8.39 & 23.82 & 27.11 & 6.05 & 19.67 \\
		\hline
		\multicolumn{7}{@{}c@{}}{\textit{using human seed words}} \\
		\hline
		\textsc{QT}  & 29.43 & 8.45 & 22.37 & 23.99 & 4.36 & 16.61 \\
		\textsc{AceSum} & 31.80 & 9.53 & 25.03 & 27.55 & 6.44 & 20.16 \\
		\thickhline
	\end{tabular}%
	\caption{ROUGE scores of \textsc{QT} and \textsc{AceSum} for aspect-specific
		summarization.}% using different kinds of seed words.}
	\label{tab:auto-seeds-results}%
\end{table}%

\subsection{Extensions to \textsc{Oposum} Dataset}

In this section, we present our additions to the \textsc{Oposum}
dataset \cite{angelidis2018summarizing}.  Firstly, we increased the
size of the review corpus. The original dataset includes only 359K
reviews, which is the result of down-sampling the \textit{Amazon
  Product Dataset} introduced in \citet{mcauley2015image}. We instead
gathered all reviews tagged with at least one of the \textsc{Oposum}
domains (``Laptop Bags'', {``Bluetooth Headsets''}, {``Boots''},
{``Keyboards''}, {``Televisions''}, and {``Vacuums''}) from the newest
version of the \textit{Amazon Product Dataset} compiled by
\citet{ni2019justifying}. Since {``Laptop Bags''} and {``Bluetooth
  Headsets''} were significantly smaller than the other four domains,
we additionally included all reviews tagged with ``{Bags}'' and
``{Headsets}''. We were able to increase the dataset to~4.13M reviews,
i.e.,~by a factor of~12.

Secondly, we created a large collection of human-written abstractive
summaries for aspect-specific summarization evaluation. For each test
product (e.g., television set) and for each aspect (e.g.,
\aspect{image quality}), we asked three annotators to write an opinion
summary about the given aspect. The annotators were shown 10 input
reviews, in which opinions about the target aspect were highlighted to
aid them in their task.  We only used the three most common aspects
for each domain, since opinions about less common aspects do not
appear frequently in reviews.  We gathered 540~aspect-specific
summaries in total.

\subsection{Example Summaries}

Finally, we show general and aspect-specific summaries produced by
\textsc{QT}, \textsc{T5-similar}, \textsc{AceSum}, and \textsc{Human} on \textsc{Space}
(Table~\ref{tab:space-example}) and \textsc{Oposum+}
(Table~\ref{tab:oposum-example}). We also show two-aspect summaries
produced by \textsc{QT} and \textsc{AceSum} in Table~\ref{tab:two-aspect-examples}.

\begin{table*}[t]
	\small
	\centering
	\renewcommand{\arraystretch}{1.2}
	\begin{tabular}{@{~}p{15.7cm}@{~}}
		\thickhline
		\multicolumn{1}{c}{\textsc{Human} summaries} \\
		\hline
		\textbf{\aspect{General}} \quad
		Staff was service focused and very welcoming. Common areas of the hotel smelled fresh because of how clean everything was. The rooms were comfortable and came with a fridge and a microwave. Food, both hot and cold, was very well presented and fresh. The hotel was located within walking distance to the French quarter and felt very safe at night.
		\\ \hline
		\textbf{\aspect{Building}} \quad
		It's older, looking at the hotel and lobby, but has lots of charm \& character.
		\\\hline
		\textbf{\aspect{Cleanliness}} \quad
		The hotel's lounge, bathrooms, hallways, and even the bedding were all clean and even smelled fresh.\\\hline
		\textbf{\aspect{Food}} \quad
		The breakfast is very good and plentiful and was more than just continental, offering eggs, sausage and grits in addition to the usual waffles, cereal, and fruit.\\\hline
		\textbf{\aspect{Location}} \quad
		The location is very good, walking distance to all major sights in French quarter.\\\hline
		\textbf{\aspect{Rooms}} \quad
		The room is comfortable and equipped with just about everything anyone could need ... a refrigerator, microwave, desk, sofa, iron and ironing board, and hairdryer. The room was also spacious and the hotel was very quiet.
		\\\hline
		\textbf{\aspect{Service}} \quad
		Hotel staff were unbelievably friendly and helpful; they often went above and beyond to be accommodating.
		\\\thickhline
\multicolumn{1}{c}{} \\ \thickhline
		\multicolumn{1}{c}{\textsc{AceSum} summaries} \\
		\hline
		\textbf{\aspect{General}} \quad
		The hotel is in a great location, close to the French quarter and the market. The room was clean and comfortable. Breakfast was good, and the staff was very helpful. There is a small restaurant in the lobby.
		\\ \hline
		\textbf{\aspect{Building}} \quad
		The lobby is a bit small. The lobby area is a little bit dated, but the rooms are very comfortable.
		\\\hline
		\textbf{\aspect{Cleanliness}} \quad
		The room was clean and comfortable. The bathroom was very clean with a nice shower.\\\hline
		\textbf{\aspect{Food}} \quad
		The breakfast was very good, with a variety of choices. The breakfast buffet was good.\\\hline
		\textbf{\aspect{Location}} \quad
		The location is great, right in the heart of Bourbon street, and within walking distance of the French quarter.\\\hline
		\textbf{\aspect{Rooms}} \quad
		The room was very spacious and the bathroom was very nice. The room had a TV, a microwave, and a separate shower. There was a small fridge in the room, which was nice.
		\\\hline
		\textbf{\aspect{Service}} \quad
		The staff was very friendly and helpful.
		\\\thickhline
\multicolumn{1}{c}{} \\ \thickhline
		\multicolumn{1}{c}{\textsc{T5-similar} summaries} \\
		\hline
		\textbf{\aspect{General}} \quad
		I stayed at the hotel for 3 nights. The room was very clean, the staff was friendly and the breakfast was excellent! The location was great - a short walk to the Eiffel Tower and St Marks Square.
		\\ \hline
		\textbf{\aspect{Building}} \quad
		i stayed here for 3 nights. The staff was very friendly and helpful, the rooms were clean and the location was great! The breakfast was good and there was a lot to do in the city.
		\\\hline
		\textbf{\aspect{Cleanliness}} \quad
		I stayed at the hotel for 3 nights. The staff was very friendly and helpful, the rooms were clean and the breakfast was excellent! The location is great - just a few minutes walk from the Eiffel Tower and it's close to the metro and Bourbon Street. \\\hline
		\textbf{\aspect{Food}} \quad
		I stayed at the San Diego for a week. The staff was very friendly and helpful, the rooms were clean and the location was great! The breakfast was good and there wasn't much to do in the room.\\\hline
		\textbf{\aspect{Location}} \quad
		I stayed at the hotel for 3 nights. The room was very clean, the staff were very helpful and the breakfast was excellent! The location is great - a short walk to the Eiffel Tower and St Marks Square. \\\hline
		\textbf{\aspect{Rooms}} \quad
		I stayed at the hotel for 3 nights. The room was very clean, the staff were very helpful and the location was great! The rooms were clean and well appointed - the breakfast was good and there was a great selection of food and drink options in the morning.
		\\\hline
		\textbf{\aspect{Service}} \quad
		I stayed at the hotel for 3 nights. The room was very clean, the staff was friendly and the breakfast was good! The location was great - a short walk to the Eiffel Tower and St Marks Square.
		\\\thickhline
\multicolumn{1}{c}{} \\ \thickhline
		\multicolumn{1}{c}{\textsc{QT} summaries} \\
		\hline
		\textbf{\aspect{General}} \quad
		Great location. The breakfast was very good. We would definitely stay here again. Room was clean. This hotel is great. The room was large with two queen beds. Nice hotel in a nice location. This is a multi-year award winning hotel. Staff were very helpful. The hotel is very clean. Front desk was friendly and helpful. The room was clean and comfy. The breakfast was average. It is very good. We enjoyed our stay here.
		\\ \hline
		\textbf{\aspect{Building}} \quad
		Plus all these fancy hotels have the irritating routine of charging around \$16 for internet access. The bad: the hotel is quite old and needs renovating.
		\\\hline
		\textbf{\aspect{Cleanliness}} \quad
		Pick this one. Toom was clean. The hotel is very clean. Great 5 star service. Room was nice and clean. This one was by far the best.\\\hline
		\textbf{\aspect{Food}} \quad
		The breakfast was very good. When you factor in the delicious complimentary breakfast consisting of scrambled eggs, grits, freshly-made waffles, bagels, bacon, sausage, cereal, toast, juice, and coffee. \\\hline
		\textbf{\aspect{Location}} \quad
		But it is just far enough away from the craziness of Bourbon and Canal streets. Walk. The location was also nice. The location of the hotel was excellent.\\\hline
		\textbf{\aspect{Rooms}} \quad
		The room was great. The bed was comfortable. The room was large with two queen beds. Hotel room was clean and comfortable.
		\\\hline
		\textbf{\aspect{Service}} \quad
		The staff was very nice. Every member of the staff we encountered was gracious, friendly and helpful. The staff at the hotel are super nice and attentive.
		\\
		\thickhline
	\end{tabular}
	\caption{\emph{General} and \emph{aspect-specific} summaries for a hotel  generated
          by four  systems (\textsc{Space} dataset).}
	\label{tab:space-example}%
\end{table*}
\begin{table*}[t]
\small
\centering
\renewcommand{\arraystretch}{1.3}
\begin{tabular}{@{~}p{15.7cm}@{~}}
	\thickhline
	\multicolumn{1}{c}{\textsc{Human} summaries} \\
	\hline
	\textbf{\aspect{General}} \quad
	The price is great. Lightweight and comfortable fit in the ear. Based on feedback from others I can be heard clearly. Very easy to use and compatible with all of my phones! It holds a charge great, is light enough. The sound quality is great, but cheap.
	\\ \hline
	\textbf{\aspect{Comfort}} \quad
	This unit is very lightweight and comfortable to fit in the ear. Makes it really easy to keep in there. Fits perfectly.
	\\\hline
	\textbf{\aspect{Ease of Use}} \quad
	The headphones are very easy to use and fit almost any cell phone. The only problem is that sometimes it is difficult to know if the equipment is on or off.\\\hline
	\textbf{\aspect{Sound Quality}} \quad
	The sound from the headphones is very good, the audio quality
        is excellent. Despite this they could have a louder sound for
        their maximum volume.		\\\thickhline
\multicolumn{1}{c}{} \\ \thickhline
	\multicolumn{1}{c}{\textsc{AceSum} summaries} \\
	\hline
	\textbf{\aspect{General}} \quad
	I've had this for a few months now and it's still working great. I haven't had any issues with the sound quality, and I am very happy with the quality of the product. The earpiece is very comfortable and is very lightweight.
	\\ \hline
	\textbf{\aspect{Comfort}} \quad
	It's a comfortable fit. The earpiece isn't too big, but it fits your ear well. It is a little snug, but my ear is very comfortable. It is comfortable to wear for long periods of time.
	\\\hline
	\textbf{\aspect{Ease of Use}} \quad
	It is easy to use, easy to set up, and easy to connect to my phone. It's not that easy to get on and off.\\\hline
	\textbf{\aspect{Sound Quality}} \quad
	The sound quality is good, but I'm not a fan. I haven't had any issues with the sound of the headphones.
		\\\thickhline
\multicolumn{1}{c}{} \\ \thickhline
	\multicolumn{1}{c}{\textsc{T5-similar} summaries} \\
	\hline
	\textbf{\aspect{General}} \quad
	I bought this headset for my son. He loves it, but a lot of people have complained about the sound quality of the headset! It is very comfortable and the mic isn't loud enough to hear. It has sounded great.
	\\ \hline
	\textbf{\aspect{Comfort}} \quad
	I bought this headset for my husband. He loves it, and it works great! It has a great sound and the sound quality is excellent - the only thing is that the microphone isn't very loud.
	\\\hline
	\textbf{\aspect{Ease of Use}} \quad
	I bought this headset for my husband. He loves it, and it is very comfortable! If you are looking for a good headset, this is the best headset you can buy for the price\\\hline
	\textbf{\aspect{Sound Quality}} \quad
	I bought these for my husband. He loves them, and they are very comfortable! They don't have a lot of noise. If you are looking for something that will work for you, then they're ok but they will not work with the earbuds.\\\hline\hline
	\multicolumn{1}{c}{\textsc{QT} summaries} \\
	\hline
	\textbf{\aspect{General}} \quad
	Thank you! The battery life is ... bizarre. Light to the ear. I highly recommend this bluetooth headset. Lightweight and comfortable fit in the ear. I returned it and received a refund. I used it mostly in my car on my commute to work. Great product.
	\\ \hline
	\textbf{\aspect{Comfort}} \quad
	I would really like it if it would stay in my ear or if the loop that went around my ear would hold it to my ear. I could not get this headset to work.
	\\\hline
	\textbf{\aspect{Ease of Use}} \quad
	Item delivery just as described! Its made of the cheapest of materials and the bluetooth has a hard time staying connected. My only gripe is that sometimes there's a small lapse between my voice. \\\hline
	\textbf{\aspect{Sound Quality}} \quad
	Also they are comfy and stay in my ears. The headset is light and fits comfortably in my ears (though it takes some time to find the right angle and fit it right in).\\
	\thickhline
\end{tabular}
\caption{\emph{General} and \emph{aspect-specific} summaries for the ``Bluetooth Headsets'' domain) generated
  by four different systems (\textsc{Oposum+} dataset).}
\label{tab:oposum-example}%
\end{table*}
\begin{table*}[t]
\small
\centering
\renewcommand{\arraystretch}{1.3}
\begin{tabular}{@{~}p{13.6cm}@{~}}
	\thickhline
	\multicolumn{1}{c}{\textsc{AceSum} summaries} \\
	\hline
	\textbf{\aspect{Cleanliness}} and \textbf{\aspect{Location}} of a hotel \\
	The hotel is clean and the rooms are very clean. The location is great, right on the beach, and close to the Eiffel Tower.
	\\ \hline
	\textbf{\aspect{Building}} and \textbf{\aspect{Cleanliness}} of a hotel \\
	The room was very clean and the bathroom was very clean. The pool was nice, but the pool area was a bit small.
	\\\hline
	\textbf{\aspect{Food}} and \textbf{\aspect{Rooms}} of a hotel \\
	The breakfast was good, the food was good and the staff was
        very friendly. The breakfast buffet was good with a variety of
        choices.\\ \hline 
	\textbf{\aspect{Quality}} and \textbf{\aspect{Size}} of a laptop bag \\
	It's a good size for a laptop. It is not a heavy bag, it is made of a soft material. \\\hline
	\textbf{\aspect{Ease of Use}} and \textbf{\aspect{Suction Power}} of a vacuum \\
	I've had this vacuum for a few months now and it's very easy to use. I don't like the fact that it is a little heavy, but it does a great job of picking up the hair. \\\hline
	\textbf{\aspect{Comfort}} and \textbf{\aspect{Looks}} of a pair of boots \\
	They are a little tight, and they are not comfortable. They look great with jeans and skirts. If you are looking for a comfortable shoe that will last a long time, do not order this. \\	\thickhline 
\multicolumn{1}{c}{} \\ \thickhline
	\multicolumn{1}{c}{\textsc{QT} summaries} \\
	\hline
	\textbf{\aspect{Cleanliness}} and \textbf{\aspect{Location}} of a hotel \\
	Overall we had a nice stay at the hotel. It's well worth the extra money. For the price I paid it underwhelmed (\$350 for 1 night). Doesn't get more LA than this have a drink at the roof top.
	\\ \hline
	\textbf{\aspect{Building}} and \textbf{\aspect{Cleanliness}} of a hotel \\
	The service was great! The hotel was beautiful. amazing. Holy cow. I love staying at this hotel. Excellent. Superb service!! I can't say enough about how perfect this hotel was for us. I stayed at this hotel not too long.
	\\\hline
	\textbf{\aspect{Food}} and \textbf{\aspect{Rooms}} of a hotel \\
	(Note that breakfast isn't necessarily included in the price.) On the first floor there is a small breakfast room but no restaurant. Also a small but cosy terrace with swimming pool. Rooms are a decent size but walls are paper thin.\\\hline
	\textbf{\aspect{Quality}} and \textbf{\aspect{Size}} of a laptop bag \\
	The hand straps have not ripped or torn so really I think the problem was that I put too much weight in the bag. Barely fit a 14 inch HP sleek notebook. I would not recommend this bag \\\hline
	\textbf{\aspect{Ease of Use}} and \textbf{\aspect{Suction Power}} of a vacuum \\
	I even tried putting ear plugs in to vacuum with it, but it still hurts my ears. I looked at every small but powerful vacuum I could find in stores and on line. \\\hline
	\textbf{\aspect{Comfort}} and \textbf{\aspect{Looks}} of a pair of boots \\
	Once the weather got cold the shoes became more stiff and they really hurt now so it looks like I wasted \$40. I am wondering if they are worth returning or just passing off to someone \\
	\thickhline
\end{tabular}
\caption{Opinion summaries focusing on \emph{two aspects} (\textsc{Space} and
  \textsc{Oposum+} datasets)}
\label{tab:two-aspect-examples}%
\end{table*}
	
\end{document}